\documentclass[runningheads]{llncs}
\usepackage{graphicx}
\usepackage{hyperref}
\usepackage{amsfonts,amssymb,amsmath}

\usepackage{xcolor, soul}
\sethlcolor{yellow}

\usepackage{nameref}
\usepackage{enumitem}   
\usepackage{verbatim}

\begin{document}
\title{Day-ahead time series forecasting: \\application to capacity planning}
\titlerunning{Day-ahead time series forecasting}
%
\author{Colin Leverger\inst{1,2}, Vincent Lemaire\inst{1}\\ Simon Malinowski\inst{2}, Thomas Guyet\inst{3}, Laurence Roz\'e\inst{4}}
%
\authorrunning{C. Leverger et al.}
%
\institute{Orange Labs, 35000 Rennes -- France \and Univ. Rennes INRIA/CNRS
IRISA, 35000 Rennes -- France \and AGROCAMPUS-OUEST/IRISA - UMR 6074, 35000
Rennes -- France, \and INSA Rennes, IRISA, 35000 Rennes -- France \\
\email{\{colin.leverger\}@orange.com}
}
\maketitle              
\begin{abstract}
In the context of capacity planning, forecasting the evolution of informatics
servers usage enables companies to better manage their computational resources.
We address this problem by collecting key indicator time series and propose to
forecast their evolution a day-ahead. Our method assumes that data is structured
by a daily seasonality, but also that there is typical evolution of indicators
within a day. Then, it uses the combination of a clustering algorithm and Markov
Models to produce day-ahead forecasts. Our experiments on real datasets show
that the data satisfies our assumption and that, in the case study, our method
outperforms classical approaches (AR, Holt-Winters).

\keywords{Time series, Capacity Planning, Clustering, Markov Models.}
\end{abstract}

\section{Introduction}
Capacity planning (CP) is one major preoccupation for today’s companies. This
ensemble of practices characterises the way an enterprise manages, updates, adds
or removes physical/virtualised servers from its infrastructure. A well-made CP
helps to reduce operational costs, and improves the quality of the provided
services: indeed, one of the major goals is to maintain the best quality of
services for the end users. Most of the time, project owners, managers and
experts apply CP guidelines to manage their infrastructure manually. This
approach is cumbersome and overestimate needs to prevent from any business
interruptions. As the manager will have to daily take decisions about its
infrastructure, we are interested in forecasting the full time series of the day
ahead. Contrary to a lot of time series forecasting techniques, the forecasting
horizon is of several steps in this study. We could then rely on the circadian
rhythm of the data to improve forecasts. 

Bodik et \textit{al.} \cite{bodik2009automatic} show that data analytics on
datacenters key performance indicators (KPI) may be used to do CP
(\textit{e.g.}, CPU, number of users, RAM). Indeed, the data generated by
servers or user activity could be valuable sources of information. One
particular data analytic task that may help resource managers in the daily
activity is to forecast the evolution of the KPI in time. That is basically a
time series forecasting task. The more accurate are forecasted the KPI, the more
informed will be the management decisions regarding CP. In particular, we make
the assumption, denoted $(\mathcal{A})$, that \textit{KPI time series are driven
by a two-order temporal process}. The circadian scale drives the daily evolution
of the KPI but, at the second order, this daily behaviour is itself driven by
some hidden rules. The evolution within a week could be considered. During the
weekdays (from Monday to Friday) KPIs have a daily behaviour which is different
from the weekend daily behaviour.

The method we propose is to capture the daily behaviour of the KPI by defining
some ``types of days'' and thus to detect second-order behaviours by analysing
sequences of typical days. For now and as a primary study, we did not take into
account the position of a day in a week (\emph{e.g.}, Mondays, Tuesdays, ...)
and only consider types of days regarding shapes of the time series data only.
The experiments compare the forecasting performance of our method with different
baseline in order to validate our hypothesis $\mathcal{A}$ on a large collection
of real data coming from a very large infrastructure (large time depth and great
variability of KPI).

\section{Day-ahead time series forecasting}

\begin{figure}[tb]
    \begin{center}
        \includegraphics[width=1\textwidth]{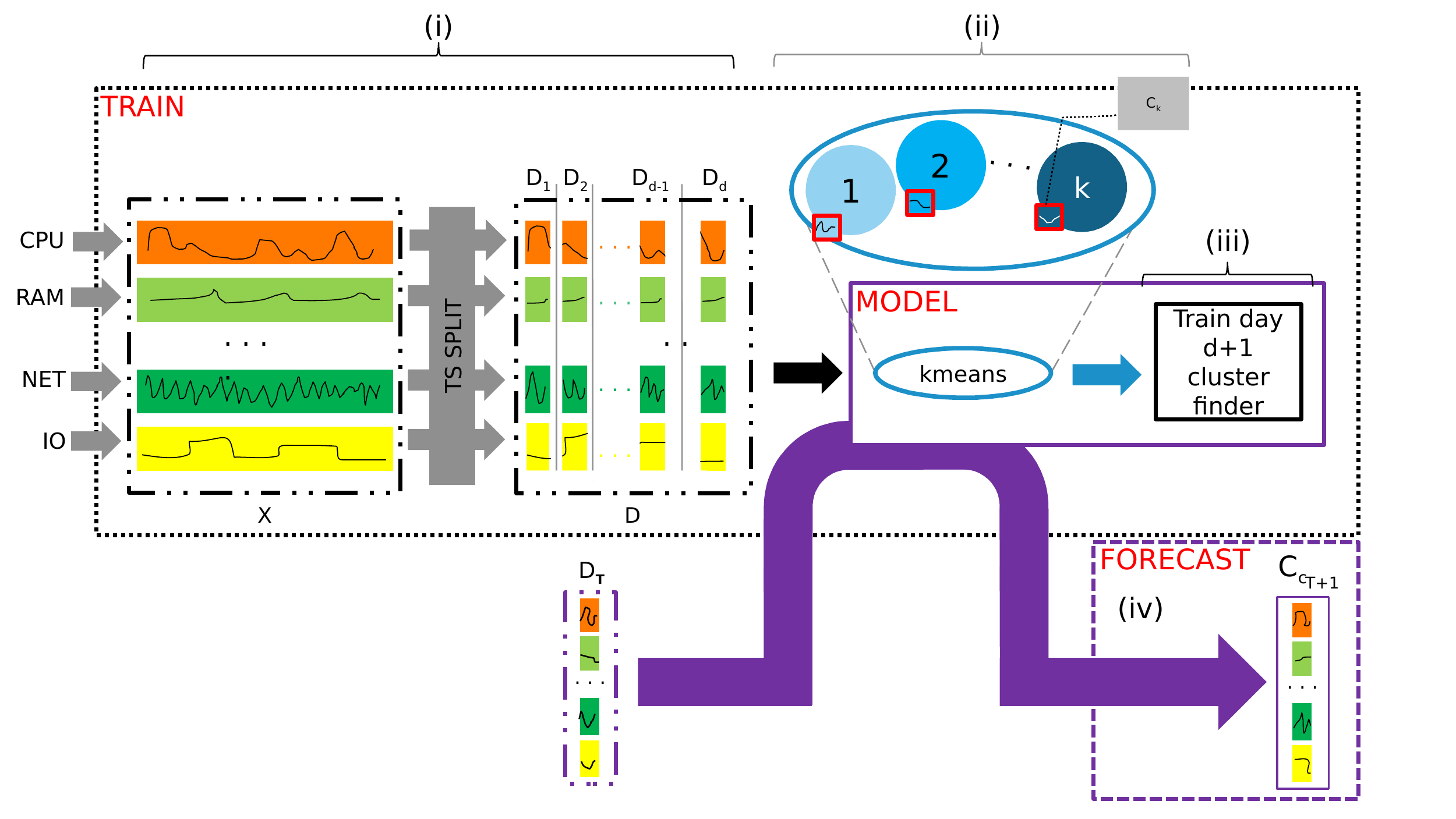}
        \caption{Illustration of the main steps of the day-ahead time series forecasting: (i) multivariate time-series splitting in daily time series, (ii) time series clustering, (iii) day sequentiality modelling, (iv) forecast on a new daily time series (see details in the text).}
    \label{fig1} 
\end{center}
\end{figure}

In this section, we present our approach to produce one-day-ahead time series
forecasting. This approach is composed of three learning steps (i) data
normalisation and split, (ii) clustering, (iii) next-day cluster estimation, and
a forecasting step (iv) next day forecasting. The learning steps take as input a
multivariate time series and build a forecasting model. In the forecasting step,
the model is applied to the time series current day in order to forecast the
next day. Fig. \ref{fig1} depicts the learning steps of our approach,
\emph{i.e.} from (i) to (iii), together with the forecasting step (iv). The four
steps are presented below.

Let $\vec{X}=\langle \vec{X}_1,\dots, \vec{X}_n\rangle$ be a multivariate time
series of length $n$.  For all $i\in[1,n]$, $\vec{X}_i\in \mathbb{R}^p$, where
$p$ denotes the dimensionality of $\vec{X}$ (the number of monitored KPI in our
application case). Let $h$ be the forecasting horizon that corresponds to one
day. 

\noindent \textbf{(i) Data normalisation and split:} First, we normalise each
dimension of $\vec{X}$ using a $\mathcal{N}(0,1)$ normalization. Then, we
construct the set $\mathcal{D} = \{\vec{D}_1, \dots, \vec{D}_d \}$, where
$\vec{D}_i$ represents the subseries of $\vec{X}$ corresponding to the $i^{th}$
day of measurements, and $d=\frac{n}{h}$. For sake of simplicity, we assume that
$n$ is a multiple of $h$.

\noindent \textbf{(ii) Clustering:} The elements of $\mathcal{D}$ are then given
to a clustering algorithm. In our case, we use the multidimensional $k$-means
based on the Euclidean distance. The centroids of the created clusters are
denoted $\vec{C}_1, \dots, \vec{C}_k$. They correspond to typical day evolution.
The choice of the number $k$ of clusters will be discussed in
Section~\ref{sec:xp}.

\noindent \textbf{(iii) Next-day cluster estimation:} The set $\mathcal{D}$ is
encoded into a sequence $\langle c_1, \dots, c_d\rangle$, where $ c_i \in [1,k]$
corresponds to the index of the cluster that contains the day $\vec{D}_i$. We
model this sequence by a first-order Markov Model of transition matrix $\Pi =
(\pi_{i,j})$, where $\pi_{i,j}$ is the probability that the next day belongs to
cluster $j$ given that the current day belongs to cluster $i$. These
probabilities are estimated using the sequence $\langle c_1, \dots, c_d\rangle$.
In other words, this Markov Model enables us to estimate the most probable
cluster to which will belong the next day of measurements.

\noindent \textbf{(iv) Next-day forecasting:} This forecasting step makes use of
the model learned above to predict the next day measurements ($\vec{D}_{T+1}$)
given the current day measurements ($\vec{D}_T$). First, the index of the
closest cluster to $\vec{D}_T$ is computed, and denoted $c_T$. The most probable
cluster for the next day is estimated using the transition matrix $\displaystyle
\Pi : c_{T+1} = \arg \max_{j \in 1, \dots, k} \pi_{c_T, j}$. Finally, the
forecasting of the next day is given by the centroid $\vec{C}_{c_{T+1}}$.

\section{Experiments and Results}
\label{sec:xp}
In this section, we briefly introduce our real capacity planning dataset. Before
the results, we present the evaluation protocol of our approach and its
underlying assumption stating time series with two-orders temporal scales.

\subsection{Data} 
This work has been endorsed by a growing project at Orange: Orange Money
(OM)\footnote{\url{https://orangemoney.orange.fr/}}. This latest was established
in late 2008. It aims in providing an easy access to bank transfers for African
customers. The bank exchange system is hosts by a large Orange infrastructure
(162 servers) and is used by more than 31 million customers across all of Africa
and Europe.

The datasets are collected from this infrastructure using Nagios supervisor. It
collects technical and operational metrics every five minutes:
\begin{itemize}
    \item \emph{Technical data}: data concerning the server’s performance such
    as percentages of CPU/memory use across all servers,
    \item \emph{Operational data}: data concerning users browsing the service;
    number of financial transactions every minute, number of people on a
    website, etc. 
\end{itemize}

\subsection{Protocol}
This study has been conducted in two major steps: 

(i) We have first experimented using univariate time series, where the day
\(T+1\) is forecasted using only past data from current time series. Our
multivariate time series have therefore first being exploited as univariate ones
(we took each features independently). 

(ii) We have also experimented using multivariate time series, where the past of
several series is used to forecast future of one KPI. We were curious about the
possible improvement of the accuracy using multivariate time series. Indeed, we
can assume that having more information describing the datacenter behaviour may
help improve forecasts.

The performance of our method is compared to four alternative approaches. Two
baseline methods are used to evaluate our assumption $\mathcal{A}$ on the time
series structure in the dataset:
\begin{itemize}
\item \textit{Mean day} is a simple mean day calculation: all days present in
  learning ensemble $\mathcal{D}$ are used to compute the average day. This
  latest is then given as a forecast result for day \(T+1\). Intuitively, it
  should give the worst forecasts if there are actually very different types of
  days.
\item \textit{Omniscient algorithm} is an adaptation of the day-ahead
  forecasting method (with same clusters) with an omniscient prediction of
  \(c_{T+1}\) instead of the Markov model. This baseline is used to evaluate the
  Markov Model independently of the clustering step. It assumes that we know to
  which cluster belong day $T +1$. The forecasts given by this method cannot be
  worse (in terms of prediction error) than the ones given by our approach.
\end{itemize}
The relative performance of our approach with regard to these two baselines
gives indication about the compliance of data with our assumption $\mathcal{A}$.
Moreover, two classical forecasting techniques are used to compare the quality
of the forecasts given by our approach: 
\begin{enumerate}
\item \textit{Auto-Regressive (AR)} model \cite{akaike1969fitting}, which is a
representation of a random process that can be used to describe some time
series, 
\item \textit{Holt Winters (HW)} Triple Exponential Smoothing
\cite{winters1960forecasting}, which is a rule of thumb technique for smoothing
time series data. It extends the Holt’s method to capture seasonality.
\end{enumerate}



These two methods have been preferred to more complex models such as ARIMA
models \cite{doi:10.1080/01621459.1982.10477767} or LSTM
\cite{doi:10.1162/neco.1997.9.8.1735} whose parameters are uneasy to configure.

We quantify the prediction error by computing the Mean Square Error (MSE)
between the forecasted values and the real values. 
The dataset is split chronologically into three parts: 70\% of the data as
training data, 15\% as validation data and 15\% as test data. The validation
dataset is used to select the best number of clusters $k$ (ranging from $2$ to
$200$) based on the MSE.

\subsection{Results}
\label{rad}

Table \ref{tab:univariate_results} presents the results obtained for 458
univariate time series (both operational and technical coming from 162 servers)
of the OM projects. To obtain more significant results, we use all day present
on the test ensemble to produce forecasts (\emph{i.e.} if the test ensemble is
composed of 6 days, we produce 5 forecasts (excluding the first day). The mean
errors are computed as the average MSE per univariate time series. For each
forecast, the methods are ranked by decreasing error, and we compute the mean
rank. 

\begin{table}[tb]
\centering
\caption{Errors and ranks for univariate forecasting results.}
\label{tab:univariate_results}
\setlength{\tabcolsep}{5pt}
\begin{tabular}{cccc}
\hline 
Algorithm & Mean error $\pm$ std & Mean rank $\pm$ std \\ 
\hline
Omniscient algorithm     & $\mathbf{0.45 \pm 0.83}$ & $\mathbf{1.51 \pm 0.79}$
\\

Day-ahead forecasting     & $0.52 \pm 0.87$          & $2.62 \pm 0.91$ \\

AR model                 & $0.73 \pm 0.92$          & $3.30 \pm 1.38$ \\

HW model                 & $\mathit{314 \pm 1775}$  & $\mathit{3.91 \pm 1.37}$ \\ 

Mean day                 & $0.71 \pm 1.03$          & $3.66 \pm 0.95$ \\ 
\hline 
\end{tabular} 
\end{table}

Let us first analyse the results to assess our assumption $\mathcal{A}$
about the underlying data structure. We observe that forecasting the centroid of
the known cluster for day $T+1$ (omniscient method) outperforms significantly
the daily mean of the whole time series. This means that our assumption about
the two temporal scales is satisfied by our dataset. The average number of
clusters selected on the validation sets is $16$ for $365$ days of data. This
number is sufficiently low to conclude that there are actually clusters of
typical days.

Unsurprisingly, the omniscient method outperforms our method as the former
always make the best choice. It also shows that the crucial step of forecasting
the type of the day ahead could be improved. In fact, Markov models are very
simple models and more advanced approaches could probably help improve our
overall approach. But we mainly notice that our algorithm has better performance
than AR and HW that is often chosen for seasonal time series forecasting. It can
be explained by a weak adequacy of these methods to our task. AR and HW are more
used to forecast next few points of the time series while our task requires to
forecast the time series for the entire day (96 points). The mean error is
higher for AR than the mean-day forecast, but its ranking is better, meaning
that AR is better than a mean day on most of the time series, but when it fails,
it fails with higher errors.

One noticeable thing is the pretty bad \textit{performances of the HW model}.
This could be explained because HW is less efficient in forecasting several
points in the future, and often base its forecasts following the last tendency
observed. 

\medskip

The previous experiments evaluate our algorithm on univariate time series. Table
\ref{tab:mres} presents  results on multivariate time series with two CPUs from
two different servers.

The multivariate mean algorithm is slightly better than our algorithm in terms
of rank but not in terms of MSE. It indicates that having more features involved
in the process increases the chance that the mean day is more representative for
the forecast, and thus better. 

One possible explanation is that the number of clusters is too small to extract
meaningful groups of days. With multivariate time series, the number of types of
day increases but we did not increase the maximum number of cluster (tested
during on the validation set). 
%
Finally, one noticeable performance of our multivariate algorithms is its low
mean error. Nonetheless, experiments on more time series are required to
conclude, 
and but it shows that the method is suitable regarding technical CPU time
series.

\begin{table}[tb]
\centering
\caption{Errors and ranks for multivariate forecasting results.
never had the lower MSE not presented.}
\label{tab:mres}
\setlength{\tabcolsep}{5pt}
\begin{tabular}{cccc}
\hline 
Algorithm & Mean error $\pm$ std & Mean rank $\pm$ std \\ 
\hline
Multivariate mean                           & $0.1 \pm 0.14$             &
$\mathbf{2.07 \pm 1.48}$ \\
Multivariate day-ahead forecasting               & $0.044 \pm 0.1$            &
$2.24 \pm 0.86$ \\ 
Omniscient multivariate algorithm     & $\mathbf{0.041 \pm 0.09}$  & $2.44 \pm
0.70$ \\
Omniscient univariate algorithm       & $0.5 \pm 0.6$              & $4.03 \pm
0.85$ \\
Univariate day-ahead forecasting                & $0.57 \pm 0.63$            &
$4.84 \pm 0.87$ \\ 
AR model                                    & $0.76 \pm 0.7$             & $5.38
\pm 1.38$ \\ 
\hline 
\end{tabular} 
\end{table}

\subsection{Discussion}
\label{rad2}

Experimental results show that our approach performs well on our real dataset,
but we also know that it is an early proposal that has some limitations and
weaknesses.

In the first stage of our model, daily time series are clustered. At the time,
we only experimented the $k$-means algorithm with a Euclidean distance. In
practice, if detecting data peaks between 2 PM and 4 PM is something of
interest, this distance is to be privileged. But if the goal is to detect peaks
in a day without precise information about their timings, DTW
\cite{niels2004dynamic} is a better candidate. In our mind, there is not a
unique good choice but better some choices that fit the data characteristics
like Douzal-Chouakria et Amblard \cite{ChouakriaA12} suggest for a
classification task. In addition, the clustering strategy could also be
evaluated. One of the weaknesses of the current approach using a $k$-means
algorithm is the critical choice of $k$. In this study, we find the optimal
number of clusters using the validation set. We test various $k$-means sizes on
the training dataset, and select the one which helps our algorithm in having a
lower MSE. Less empirical techniques such as David et Bouldin criteria
\cite{davies1979cluster} or even Silhouette \cite{rousseeuw1987silhouettes}
could help to select a priori the best number of clusters with a lower
computational cost.

An interesting thing would be to know which of the steps described on the Figure
\ref{fig1} is lowering forecast performances. This could help us in improving
the chain by tuning very particular and identified parameters. The omniscient
algorithm that knows for sure the day $T+1$ is useful as it reveals that having
an almost perfect clustering algorithm could greatly enhance precision of
forecasts (see Tables \ref{tab:univariate_results} and \ref{tab:mres} which show
that omniscient technique often outputs forecasts with lower mean error).
Enhancing the quality of clustering could then be a key to better results.

In the second stage of our model, the experiment shows that there is possible
improvements of the prediction of the type of the next day. 
A more accurate solution than Markov models can bridge the gap with the
omniscient approach. At the time, a Markov model takes only into account the day
before the one to predict. This simple model has been preferred to higher order
Markov models because of the required quantity of training data. With $7$
different days, transition matrix of size $49$ are to learn in our case, but of
size $343$ for a 2-order Markov model. This requires long sequences of days to
accurately estimate them. We currently collected $365$ days of data history and
thus prefer to focus on a simple but sound approach. A study of state-of-the-art
approaches of sequence prediction with sparse data will help us to identify good
candidates to replace the Markov model.

Finally, as mentioned in the introduction, our model does not take into account
the position of the weekday to make forecasts. However, some observations
let us believe that it could improve the prediction accuracy. In fact, we
noticed that the clustering of daily time series set up with 2 clusters will
extract two types of behaviours: A high activity profile (HA) from Mondays to
Fridays and a low activity profile (LA) on Saturdays and Sundays. As a
consequence, the trained Markov Model will more likely predict a HA day after
another HA day (with probability 0.8). This means that every Friday, it will
wrongly predict the day with a HA profile. A model with the weekday information
will split the next day prediction rule in two different rules: High activity
from Monday to Thursday leads to a high activity the next day while high
activity on Friday leads to a low activity. It remains interesting to have
information about the types of the days as, in real data, there are several
different of profiles of days that are not necessarily correlated to the weekday
information.

\section{Conclusion}
In this work, we presented a general method to address a specific problem of
capacity planning, \textit{i.e.} the forecast of the evolution of KPI indicators
a day ahead. The method we propose is a time series forecasting method that is
founded on the assumption that the time series are implicitly structured as a
sequence of typical days. Our experiments comparing baseline approaches and
classical time series forecasting methods to our method show that this
assumption is fitted by most of the time series of our dataset. We now have to
explore more deeply the characteristics of the data that does not fit it. 
to improve the overall approach. We observe an interestingly low mean square
error for our algorithm on multivariate CPU and find these preliminary results
promising. But the proposed method may also benefit from improvements in the two
main stages: the clustering of time series and the sequential prediction. We
strongly believe that Markov model may be improved by some more recent works on
sequence prediction \cite{budhathoki2018causal}.

%
%
%
\bibliographystyle{splncs04}
\bibliography{biblio_example}

\end{document}